\documentclass{article}
\usepackage{spconf,amsmath,graphicx}
\usepackage{amsfonts,amssymb}
\usepackage{multirow}
\usepackage{float}
\usepackage{color}
\usepackage{hyperref}
\hypersetup{
    colorlinks=true,
    urlcolor=magenta,
}
\title{Deep Double Self-expressive Subspace Clustering}
\name{Ling Zhao, Yunpeng Ma, Shanxiong Chen, Jun Zhou*\thanks{*Corresponding author. This work was financially supported by  the Natural Science Foundation of Chongqing, China (cstc2021jcyj-msxmX0066).}}
\address{College of Computer and Information Science, Southwest University, Chongqing, China}
\begin{document}
%\ninept
%
\maketitle
\begin{abstract}
Deep subspace clustering based on auto-encoder has received wide attention. However, most subspace clustering based on auto-encoder does not utilize the structural information in the self-expressive coefficient matrix, which limits the clustering performance. In this paper, we propose a double self-expressive subspace clustering algorithm. The key idea of our solution is to view the self-expressive coefficient as a feature representation of the example to get another coefficient matrix. Then, we use the two coefficient matrices to construct the affinity matrix for spectral clustering. We find that it can reduce the subspace-preserving representation error and improve connectivity. To further enhance the clustering performance, we proposed a self-supervised module based on contrastive learning, which can further improve the performance of the trained network. Experiments on several benchmark datasets demonstrate that the proposed algorithm can achieve better clustering than state-of-the-art methods.
\end{abstract}
\section{Introduction}
\label{sec:intro}
In recent years, subspace clustering \cite{vidal2011subspace} has attracted extensive attention in dealing with high-dimensional data. Subspace clustering assumes that the data in the same cluster lie in the same subspace while the data in different clusters lie in different subspaces. It is widely used in face clustering, video motion segmentation, image clustering, and other fields \cite{elhamifar2013sparse}.

The most popular subspace clustering algorithm in the past decade is the sparse subspace clustering proposed by \cite{elhamifar2013sparse}. Its main idea is that each data can be represented linearly by other data in the subspace to which it belongs. That is a "self-expressive" model. After obtaining the self-expressive coefficient matrix, an affinity matrix can be constructed, which can be used for spectral clustering to obtain the final clustering result. The common practice is to impose a regularization term on the coefficient matrix to obtain a sparse coefficient matrix and ensure that each data point is linearly represented only by data points from the same subspace  \cite{elhamifar2013sparse}. For instance, adds L1-Norm restriction to coefficient matrix. Currently, many subspace clustering algorithms have been proposed to obtain the sparse coefficient matrix \cite{lu2012robust,vidal2014low,you2016oracle}.

However, the above methods cannot deal with the situation that data is distributed in nonlinear subspaces. Because of this, some sparse subspace clustering algorithms based on kernel methods have been proposed \cite{patel2014kernel}. The performance of these algorithms depends heavily on the selected kernel functions. Since the powerful nonlinear mapping capability of neural network, \cite{ji2017deep} proposed a deep subspace clustering network based on auto-encoder. The key idea is to introduce a fully-connected layer between the encoder and decoder. After the input data passes through the encoder, it is mapped to a linear subspace. Then, it will implement the self-expressive algorithm through the fully-connected layer and use the weight of the fully-connected layer as the coefficient matrix for spectral clustering. Finally, the data will be reconstructed through the decoder. In recent years, many deep subspace clustering networks based on auto-encoder have been proposed \cite{zhou2018deep,zhang2019self,dang2020multi,kheirandishfard2020multi,lv2021pseudo,peng2022adaptive,li2021lrsc}.

In the above auto-encoder based methods, most of them are directly used for spectral clustering after obtaining the coefficient matrix, ignoring the information contained in the coefficient matrix. In fact, the coefficient matrix obtained is sparse and contains the structural information between examples. These methods did not take into account the structure information contained in the coefficient matrix, which may limit the performance of subspace clustering. \cite{xu2022linearity} proposed a linearity-aware metric based on Pearson coefficient, it regards the self-expressive coefficient as a representation of the sample and learn a similarity matrix by proposing linearity-aware metric. \cite{peng2022adaptive} proposed network simultaneously consider the attribute and structure information in an adaptive graph fusion manner. It constructs a structure matrix through the coefficient matrix and performs a ”self-expressive” on structure matrix. Finally, it fuses the two coefficient matrices through an attention mechanism and achieved good results.

In this paper, a double self-expressive deep subspace clustering network is proposed. The key idea is to view the coefficient matrix as another representation of the examples. We perform ”self-expressive” to obtain another coefficient matrix. Then, the affinity matrix for spectral clustering is constructed according to the two coefficient matrices. This “double self-expressive” module is similar to but not identical to the “structure self-expressiveness” in \cite{peng2022adaptive}. In \cite{peng2022adaptive}, a structure matrix is constructed based on the coefficient matrix, and then “self-expressive” is performed on the structure matrix. We are directly performing “self-expressive” on the coefficient matrix. In addition, in order to better training network, we introduce the contrastive learning \cite{he2020momentum,chen2020simple} method into the training process. It further improve clustering performance. 

\begin{figure}
\centering
\includegraphics[width=8cm]{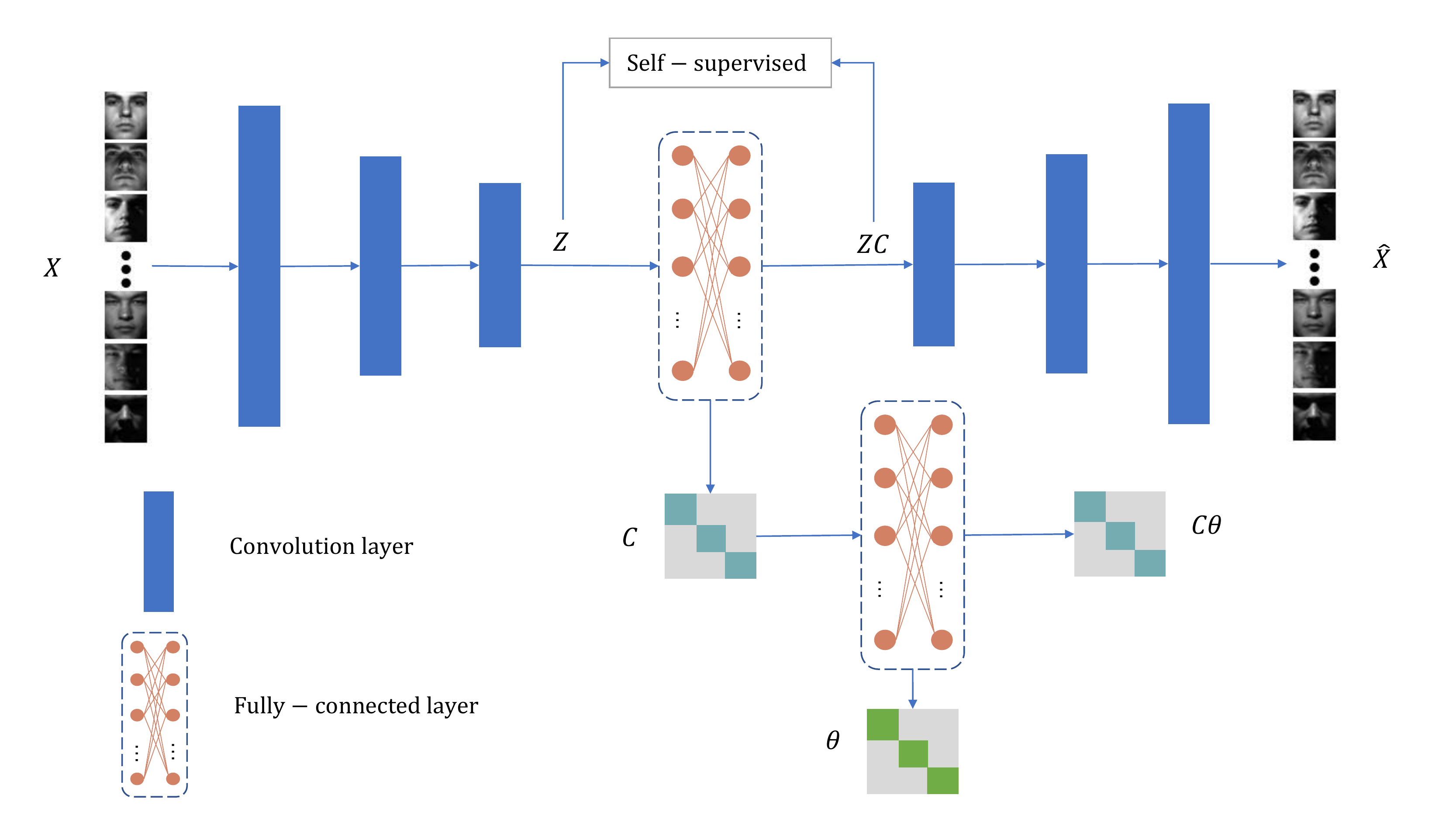}
\caption{Architecture of the proposed network. It including three section: 1) The input $X$ is mapped to $Z$ through an encoder, and $Z$ is self-expressed by $ZC$. Then, it was reconstructed as $ \hat{X}$ through a decoder. 2) The $C$ is self-expressed by $C\theta$. 3) A self-supervised module is proposed according to $Z$ and $ZC$.}
\label{fig:main}
\end{figure}

\section{method}
\label{sec:method}
In this section, we introduce the whole network structure. As shown in Fig.\ref{fig:main}, it including deep subspace clustering module, double self-expressive module, and self-supervised module.

\subsection{Deep subspace clustering}
Assuming that $X\in {\mathbb{R}}^{n \times d}$ represents $n$ examples and the dimension of each example is $d$. The "self-expressive" model represents each data point as a linear combination of all other data points:
\begin{equation}
    x_i=\sum_{j\neq i}^nx_jc_{ij}
\label{eq:1}
\end{equation}
Where $c_{ij}$ is the self-expressive coefficient. 
In order to obtain a solution with subspace-preserving, each data point is linearly represented only by data points from the same subspace. A regularization term is applied to the self-expressive coefficient matrix. Then, the sparse subspace clustering algorithm can be expressed as:
\begin{equation}
    min||C||_p  \quad\quad  s.t. \quad X = XC,\quad diag(C) = 0
\label{eq:2}
\end{equation}
Where, $p$ is the regularizer, and $diag(C) = 0$ is to avoid trivial solutions. Construct the affinity matrix according to the self-expressive coefficient matrix:
\begin{equation}
    W = \frac{|C|+|C^T|}{2}
\label{eq:3}
\end{equation}
Finally, the clustering results can be obtained by spectral clustering of the affinity matrix. The objective function of spectral clustering is:
\begin{equation}
    \underset{Q}{\min}\sum_{i,j}w_{ij}||\textbf{q}_i-\textbf{q}_j||^2_2
\label{eq:4}
\end{equation}
Where $Q$ is indicator matrix that is $Q_{ij} = 1$ if the $i$-th data belongs to the $j$-th cluster, otherwise it equals 0. In practice, since the problem is NP-Hard, spectral clustering techniques usually relax the constraint $QQ^T = I$, where $I$ is identity matrix. Finally, the final clustering result can be obtained by performing $k$-means algorithm on $Q$. 

In order to deal with complex (nonlinear) data points, \cite{ji2017deep} proposed a deep subspace clustering network (DSC-Net). Assuming that $f(\cdot)$ is an encoder and $g(\cdot)$ is a decoder, then the loss function of DSC-Net can be expressed as:
\begin{small}
\begin{equation}
    L_1 = \frac{1}{2}||X-\hat{X}||_F^2 + \lambda_1||C||_p + \lambda_2||Z - ZC||^2_F \quad s.t.\quad diag(C) = 0
\label{eq:5}
\end{equation}
\end{small}
Where, $Z = f(X), \hat{X} = g(ZC), \lambda_1, \lambda_2$ are tradeoff parameters. The network is trained by back propagation. After training, the weight of the fully-connected layer is view as a self-expressive coefficient to construct affinity matrix for spectral clustering.

\subsection{Double self-expressive}
After obtaining the self-expressive coefficient matrix, it contains the structural information between the examples. The coefficient of each example is not 0 only with the examples belonging to the same subspace. Then, the value of each cluster of example is not 0 in the same dimensions, while the value of the other dimensions is 0 or close to 0. So we think that each row in the coefficient matrix can be regarded as a representation of the example, and each cluster of example is in the same feature subspace. The self-expressive algorithm is also applicable in here. Therefore, we input the coefficient matrix as a representation of the example to another self-expressive layer (fully-connected layer), the loss function is:
\begin{equation}
    L_2 = \lambda_3||C-C\theta||_F^2+\lambda_4||\theta||_p \quad s.t.\quad diag(\theta) = 0
\label{eq:6}
\end{equation}
Where $\theta$ is the second fully-connected layer weights, $\lambda_3, \lambda_4$ is tradeoff parameters. Finally, the two coefficient matrices are fused to construct the affinity matrix:
\begin{equation}
    W = \frac{|C+\theta|+|(C+\theta)^T|}{2}
\label{eq:7}
\end{equation}
Obviously, the affinity matrix incorporates the structural information contained in the coefficient matrix.

\subsection{Self-supervised}
Inspired by \cite{he2020momentum,chen2020simple}, we introduce a self-supervised module to assist encoder network training. The main idea is to use the contrastive learning method proposed by \cite{chen2020simple}. It defining positive and negative example pairs and enlarged the similarity between positive example pairs. Then, we take $z_i$ and $z_ic_i$ as positive example pairs, and the loss function is:
\begin{equation}
    L_3 = -\sum_{i=1}^{n}log\frac{exp(cos(z_i,z_ic_i)/\tau)}{\sum_{j=1}^n exp(cos(z_i,z_ic_i)/\tau)}
\label{eq:8}
\end{equation}
Where $\tau$ is temperature parameter, $cos(a,b)$ represent the cosine similarity between vector $a$ and vector $b$.

\subsection{Training Settings}
In summary, the loss function of the whole network is:
\begin{equation}
    L = L_1+L_2+\gamma L_3
\label{eq:9}
\end{equation}
Where $\gamma$ is the indicate the hyper-parameter of self-supervised loss. It is difficult to train the whole network directly, similar to \cite{ji2017deep,zhang2019self,lv2021pseudo},we use a two-stage strategy to train the network: 
1) Pre-train the auto-encoder without self-expressive layer to get a great initialization weight. The loss function at this stage is:
\begin{equation}
    L_0 = \frac{1}{2}||X-\hat{X}||^2_F
\label{eq:10}
\end{equation}
2) Train the whole network with all mentioned modules.

After the training, we construct the affinity matrix according to Eq.(\ref{eq:7}) and use spectral clustering to get the final clustering result.

\section{expriment}
In this section, we introduce our experiment. It mainly includes datasets, evaluation indicators, experimental results comparison with existing methods, ablation experiments, and parameter analysis.

\subsection{Datasets}
In experiment, we used four benchmark datasets, i.e., Extended Yale B, ORL, Umist, and COIL20.

\textbf{Extended Yale B}\footnote{http://vision.ucsd.edu/~leekc/ExtYaleDatabase/ExtYaleB.html} is a grayscale face image dataset. It consists of 38 subjects, each of which is represented with 64 face images acquired under different illumination conditions.% Following the experimental setup of \cite{ji2017deep}, we down-exampled the original face images from 192 × 168 to 48 × 42 pixels,

\textbf{ORL}\footnote{http://www.cl.cam.ac.uk/research/dtg/attarchive/facedatabase.html} contains 40 individuals with 400 images, in which each image is captured at different scenes.

\textbf{Umist}\footnote{http://images.ee.umist.ac.uk/danny/database.html} contains 564 images of 20 individuals, in which each individual presents a range of poses from profile to frontal views. In this paper, we used 480 images of 20 people and down-exampled each image to 32 × 32.

\textbf{COIL20}\footnote{https://www.cs.columbia.edu/CAVE/software/softlib/coil-20.php} consists of 1440 gray-scale image examples, distributed over 20 objects such as duck and car model. Each object was placed on a turntable against a black background, and 72 images were taken at pose intervals of 5 degrees. %Following \cite{ji2017deep}, we down-exampled the images to 32 × 32.
\begin{figure}
    \centering
    \includegraphics[width = 8cm]{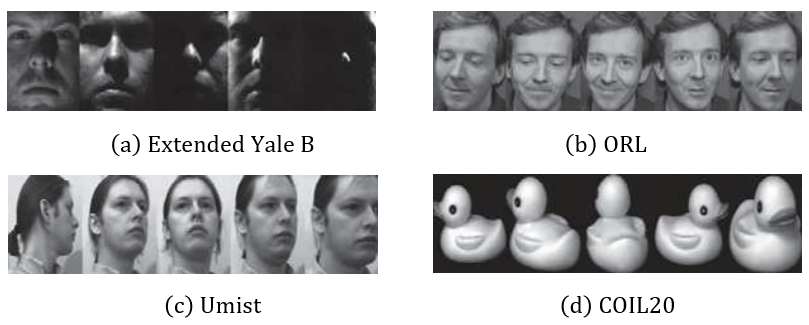}
    \caption{Example images of the four benchmark datasets.}
    \label{fig:dataset}
\end{figure}

Clustering accuracy (ACC) and normalized mutual information (NMI) were chosen as evaluation indicators. Meanwhile, we also evaluated the connectivity (CONN) \cite{chen2020stochastic} of the affinity matrix and the subspace-preserving representation error (SRE) \cite{you2016scalable} in the ablation experiment.

\subsection{Implementation Details}
Our network is implemented by PyTorch \cite{paszke2019pytorch} and optimized by Adam \cite{kingma2014adam}. For the Extended Yale B, ORL, and Umist datasets, we use three convolution layer in the encoder. Similarly, we use three transposed convolution layer in the decoder. For COIL20 dataset, we use one convolution layer in the encoder and one transposed convolution layer in the decoder. The number of convolution layer channels(c) and convolution kernel(k) parameters of each dataset are set as Table.\ref{tab:tab1}. More details can be found in \href{https://github.com/zhaoling0229/DSESC}{https://github.com/zhaoling0229/DSESC}
\begin{table}[h]
    \centering
    \caption{Convolution layer parameters for each dataset.}
    \begin{tabular}{|c|c|c|}
    \hline
        Dataset & Encoder(c/k) & Decoder(c/k) \\\hline
        Extended Yale B & \{10,20,30\}/\{5,3,3\} & \{30,20,10\}/\{3,3,5\} \\\hline
        ORL & \{5,3,3\}/\{5,3,3\} & \{3,3,5\}/\{3,3,5\} \\\hline
        Umist & \{20,10,5\}/\{5,3,3\} & \{5,10,20\} /\{3,3,5\} \\\hline
        COIL20 & \{15\}/\{3\} & \{15\}/ \{3\} \\\hline
    \end{tabular}-
    
    \label{tab:tab1}
\end{table}
\begin{table*}[h]
    \centering
    \caption{Clustering results on benchmark datasets, the best methods are highlighted in boldface.}
    \begin{tabular}{|c|c|c|c|c|c|c|c|c|}
    \hline
        \multirow{2}{*}{Method} & \multicolumn{2}{|c|}{Extended Yale B} & \multicolumn{2}{|c|}{ORL} & \multicolumn{2}{|c|}{Umist}& \multicolumn{2}{|c|}{COIL20}  \\\cline{2-9}
         & ACC & NMI & ACC & NMI & ACC & NMI & ACC & NMI \\\hline
        DSC-Net-L1 & 0.9667 & 0.9568 & 0.8550 & 0.9023 & 0.7242 & 0.7556 & 0.9314 & 0.9353\\\hline
        
        DSC-Net-L2 & 0.9733 & 0.9634 & 0.8600 & 0.9034 & 0.7312 & 0.7662 & 0.9368 & 0.9408\\\hline
        
        DASC & 0.9856 & \textbf{0.9801} & 0.8825 & 0.9315 & 0.7668 & 0.8024 & 0.9639 & 0.9686\\\hline
        
        $S^2$ConvSCN & 0.9848 & -- & 0.8950 & -- & -- & -- & 0.9786 & --\\\hline
        
        MLRDSC & \textbf{0.9864} & -- & 0.8875 & -- & -- & -- & 0.9792 & --\\\hline
        
        PSSC & -- & -- & 0.8675 & 0.9349 & 0.7917 & 0.8670 & 0.9772 & 0.9779\\\hline
        \textbf{Our} & 0.9794 & 0.9725 & \textbf{0.8975} & \textbf{0.9416} & \textbf{0.8250} & \textbf{0.8910} & \textbf{0.9833} & \textbf{0.9810}\\\hline
    \end{tabular}
    
    \label{tab:result}
\end{table*}

\subsection{Result}
In this part, we show the contrastive experiments result with other subspace clustering algorithms. The selected comparison algorithms are: DSC-Net \cite{ji2017deep}, DASC \cite{zhou2018deep}, $S^2$ConvSCN \cite{zhang2019self}, MLRDSC \cite{kheirandishfard2020multi}, PSSC \cite{lv2021pseudo}. The results on the evaluation indicators ACC and NMI are as Table \ref{tab:result}. Our method achieves state-of-the-art performance on
the datasets ORL, Umist, COIL20 and competitive results on
Extended Yale B.

\subsection{Ablation Study}
In order to verify the effectiveness of each module of the proposed algorithm, we conducted ablation experiments and the results are as Table \ref{tab:ala_1}.

\begin{figure}
    \centering
    \includegraphics[width=9cm]{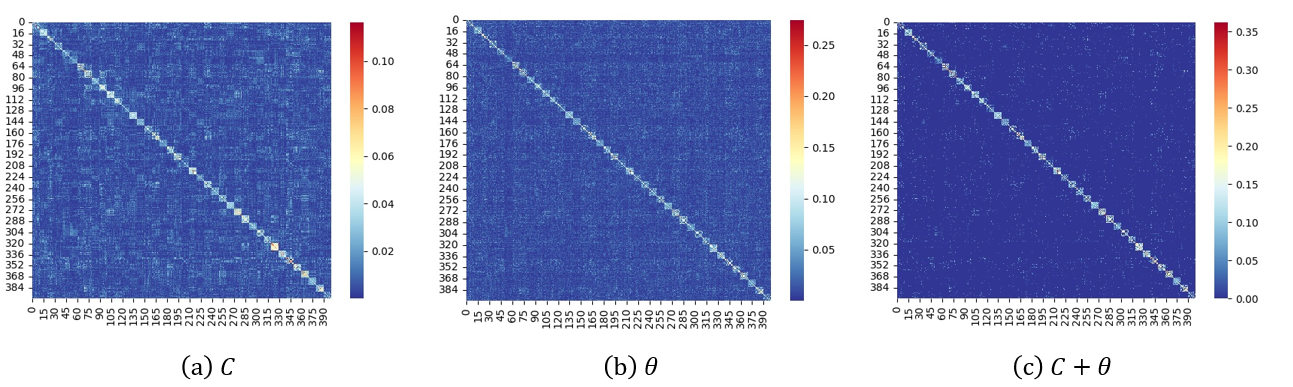}
    \caption{The visual heatmaps of affinity matrix constructed by $C, \theta, C+\theta$, respectively.}
    \label{fig:vis}
\end{figure}

At the same time, the connectivity of affinity matrix and the subspace-preserving error are recorded, and the results are as Table \ref{tab:ala_2}. To some extent, it explains why the proposed algorithm is effective. Due to the integration of structural information, the connectivity of affinity matrix is improved and the subspace-preserving representation error is reduced. Furthermore, we will visualize the affinity matrix constructed according to each module as shown in Fig.\ref{fig:vis}. It can be seen that the diagonal structure of affinity matrix constructed from two coefficient matrices is more obvious.
\begin{table}[htbp]
    \centering
    \caption{Results of ablation experiments on ORL and COIL20 datasets.}
    \begin{tabular}{|c|c|c|c|c|c|}
    \hline
     \multirow{2}{*}{Module} & \multicolumn{2}{|c|}{ORL} & \multicolumn{2}{|c|}{COIL20}\\\cline{2-5}
          & ACC & NMI & ACC & NMI\\\hline
        $L_1$ & 0.8600 & 0.9034 & 0.9368 & 0.9408\\\hline
        $L_1+L_2$ & 0.8950 & 0.9407 & 0.9667 & 0.9710 \\\hline
        $L_1+L_2+L_3$  & \textbf{0.8975} & \textbf{0.9416} & \textbf{0.9833} & \textbf{0.9810} \\\hline
    \end{tabular}
    
    \label{tab:ala_1}
\end{table}
\begin{table}[htbp]
    \centering
    \caption{The SRE, CONN of affinity matrix constructed by $C,\theta,C+\theta$, respectively.}
    \begin{tabular}{|c|c|c|c|c|}
    \hline
     \multirow{2}{*}{Module} & \multicolumn{2}{|c|}{ORL} &  \multicolumn{2}{|c|}{COIL20}  \\\cline{2-5}
         & SRE & CONN & SRE & CONN  \\\hline
        $C$ & 88.92 & 0.1755 & 81.71 & 0.2125 \\\hline
        $\theta$ & 89.15 & \textbf{0.2632} & 81.63 & 0.2231 \\\hline
        $C+\theta$ & \textbf{38.06} & 0.2531 & \textbf{81.36} & \textbf{0.4344} \\\hline
    \end{tabular}
    
    \label{tab:ala_2}
\end{table}

\subsection{parameter analysis}
\begin{figure}
    \includegraphics[width=8.5cm]{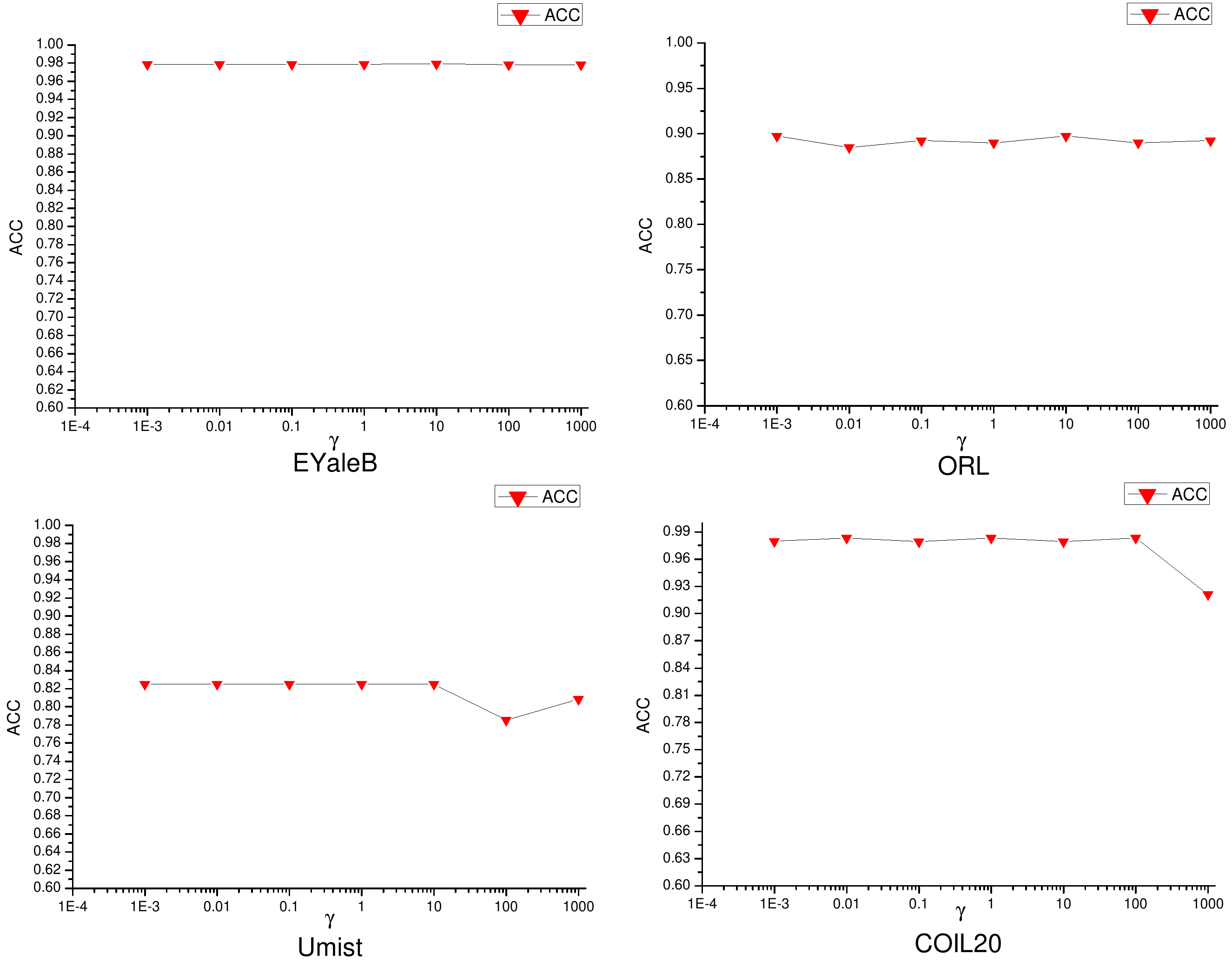}
    \caption{The influence of $\gamma$ on ACC of benchmark datasets.}
    \label{fig:parameter}
\end{figure}

We regulated the parameters of the $\gamma$ value for \{0.001, 0.01, 0.1, 1, 10, 100, 1000\} and observed the change of the ACC. The result is shown in Fig.\ref{fig:parameter}.
Although our method’s performance is influenced by this parameters, our method performs well for a wide range of values.

\section{conclusion}
We propose a deep double self-expressive subspace clustering network. The main idea is to view the self-expressive coefficient as another representation of the example and construct the affinity matrix for spectral clustering according to the two self-expressive coefficient matrices. It can improve the subspace clustering performance, and  the reasons was analyzed by subspace-preserving representation error and connectivity of the subspace. Meanwhile, in order to further improve the clustering performance, we propose a self-supervised module. Finally, experiments on several benchmark datasets demonstrate the effectiveness of the proposed algorithm.

\bibliographystyle{IEEEbib}
\bibliography{main}

\begin{thebibliography}{10}

\bibitem{vidal2011subspace}
Ren{\'e} Vidal,
\newblock ``Subspace clustering,''
\newblock {\em IEEE Signal Processing Magazine}, vol. 28, no. 2, pp. 52--68,
  2011.

\bibitem{elhamifar2013sparse}
Ehsan Elhamifar and Ren{\'e} Vidal,
\newblock ``Sparse subspace clustering: Algorithm, theory, and applications,''
\newblock {\em IEEE transactions on pattern analysis and machine intelligence},
  vol. 35, no. 11, pp. 2765--2781, 2013.

\bibitem{lu2012robust}
Can-Yi Lu, Hai Min, Zhong-Qiu Zhao, Lin Zhu, De-Shuang Huang, and Shuicheng
  Yan,
\newblock ``Robust and efficient subspace segmentation via least squares
  regression,''
\newblock in {\em European conference on computer vision}. Springer, 2012, pp.
  347--360.

\bibitem{vidal2014low}
Ren{\'e} Vidal and Paolo Favaro,
\newblock ``Low rank subspace clustering (lrsc),''
\newblock {\em Pattern Recognition Letters}, vol. 43, pp. 47--61, 2014.

\bibitem{you2016oracle}
Chong You, Chun-Guang Li, Daniel~P Robinson, and Ren{\'e} Vidal,
\newblock ``Oracle based active set algorithm for scalable elastic net subspace
  clustering,''
\newblock in {\em Proceedings of the IEEE conference on computer vision and
  pattern recognition}, 2016, pp. 3928--3937.

\bibitem{patel2014kernel}
Vishal~M Patel and Ren{\'e} Vidal,
\newblock ``Kernel sparse subspace clustering,''
\newblock in {\em 2014 ieee international conference on image processing
  (icip)}. IEEE, 2014, pp. 2849--2853.

\bibitem{ji2017deep}
Pan Ji, Tong Zhang, Hongdong Li, Mathieu Salzmann, and Ian Reid,
\newblock ``Deep subspace clustering networks,''
\newblock {\em Advances in neural information processing systems}, vol. 30,
  2017.

\bibitem{zhou2018deep}
Pan Zhou, Yunqing Hou, and Jiashi Feng,
\newblock ``Deep adversarial subspace clustering,''
\newblock in {\em Proceedings of the IEEE Conference on Computer Vision and
  Pattern Recognition}, 2018, pp. 1596--1604.

\bibitem{zhang2019self}
Junjian Zhang, Chun-Guang Li, Chong You, Xianbiao Qi, Honggang Zhang, Jun Guo,
  and Zhouchen Lin,
\newblock ``Self-supervised convolutional subspace clustering network,''
\newblock in {\em Proceedings of the IEEE/CVF conference on computer vision and
  pattern recognition}, 2019, pp. 5473--5482.

\bibitem{dang2020multi}
Zhiyuan Dang, Cheng Deng, Xu~Yang, and Heng Huang,
\newblock ``Multi-scale fusion subspace clustering using similarity
  constraint,''
\newblock in {\em Proceedings of the IEEE/CVF Conference on Computer Vision and
  Pattern Recognition}, 2020, pp. 6658--6667.

\bibitem{kheirandishfard2020multi}
Mohsen Kheirandishfard, Fariba Zohrizadeh, and Farhad Kamangar,
\newblock ``Multi-level representation learning for deep subspace clustering,''
\newblock in {\em Proceedings of the IEEE/CVF Winter Conference on Applications
  of Computer Vision}, 2020, pp. 2039--2048.

\bibitem{lv2021pseudo}
Juncheng Lv, Zhao Kang, Xiao Lu, and Zenglin Xu,
\newblock ``Pseudo-supervised deep subspace clustering,''
\newblock {\em IEEE Transactions on Image Processing}, vol. 30, pp. 5252--5263,
  2021.

\bibitem{peng2022adaptive}
Zhihao Peng, Hui Liu, Yuheng Jia, and Junhui Hou,
\newblock ``Adaptive attribute and structure subspace clustering network,''
\newblock {\em IEEE Transactions on Image Processing}, vol. 31, pp. 3430--3439,
  2022.

\bibitem{li2021lrsc}
Changsheng Li, Chen Yang, Bo~Liu, Ye~Yuan, and Guoren Wang,
\newblock ``Lrsc: learning representations for subspace clustering,''
\newblock in {\em Proceedings of the AAAI Conference on Artificial
  Intelligence}, 2021, vol.~35, pp. 8340--8348.

\bibitem{xu2022linearity}
Yesong Xu, Shuo Chen, Jun Li, and Jianjun Qian,
\newblock ``Linearity-aware subspace clustering,''
\newblock in {\em Proceedings of the AAAI Conference on Artificial
  Intelligence}, 2022, vol.~36, pp. 8770--8778.

\bibitem{he2020momentum}
Kaiming He, Haoqi Fan, Yuxin Wu, Saining Xie, and Ross Girshick,
\newblock ``Momentum contrast for unsupervised visual representation
  learning,''
\newblock in {\em Proceedings of the IEEE/CVF conference on computer vision and
  pattern recognition}, 2020, pp. 9729--9738.

\bibitem{chen2020simple}
Ting Chen, Simon Kornblith, Mohammad Norouzi, and Geoffrey Hinton,
\newblock ``A simple framework for contrastive learning of visual
  representations,''
\newblock in {\em International conference on machine learning}. PMLR, 2020,
  pp. 1597--1607.

\bibitem{chen2020stochastic}
Ying Chen, Chun-Guang Li, and Chong You,
\newblock ``Stochastic sparse subspace clustering,''
\newblock in {\em Proceedings of the IEEE/CVF conference on computer vision and
  pattern recognition}, 2020, pp. 4155--4164.

\bibitem{you2016scalable}
Chong You, Daniel Robinson, and Ren{\'e} Vidal,
\newblock ``Scalable sparse subspace clustering by orthogonal matching
  pursuit,''
\newblock in {\em Proceedings of the IEEE conference on computer vision and
  pattern recognition}, 2016, pp. 3918--3927.

\bibitem{paszke2019pytorch}
Adam Paszke, Sam Gross, Francisco Massa, Adam Lerer, James Bradbury, Gregory
  Chanan, Trevor Killeen, Zeming Lin, Natalia Gimelshein, Luca Antiga, et~al.,
\newblock ``Pytorch: An imperative style, high-performance deep learning
  library,''
\newblock {\em Advances in neural information processing systems}, vol. 32,
  2019.

\bibitem{kingma2014adam}
Diederik~P Kingma and Jimmy Ba,
\newblock ``Adam: A method for stochastic optimization,''
\newblock {\em arXiv preprint arXiv:1412.6980}, 2014.

\end{thebibliography}

\end{document}